\pdfoutput=1
\documentclass[11pt]{article}

\usepackage[preprint]{acl}

\usepackage{times}
\usepackage{latexsym}

\usepackage[T1]{fontenc}

\usepackage[utf8]{inputenc}

\usepackage{microtype}

\usepackage{inconsolata}

\usepackage{graphicx}

\usepackage{comment}
\usepackage{amsmath,amssymb,amsfonts}
\usepackage{algorithmic}
\usepackage{textcomp}
\usepackage{xcolor}
\usepackage{colortbl}  %
\usepackage{booktabs}
\usepackage{multirow}
\usepackage[most]{tcolorbox}  %
\newtcolorbox{promptbox}[1][]{enhanced, breakable, colback=gray!4, colframe=gray!45, boxrule=0.5pt, arc=2pt, left=4pt, right=4pt, top=3pt, bottom=3pt, fonttitle=\small\bfseries, coltitle=black, colbacktitle=gray!15, title={#1}, fontupper=\footnotesize}
\newtcolorbox{casebox}[1][]{enhanced, breakable, colback=white, colframe=teal!60!black, boxrule=0.6pt, arc=2pt, left=5pt, right=5pt, top=3pt, bottom=3pt, fonttitle=\small\bfseries, colbacktitle=teal!10, coltitle=black, title={#1}, fontupper=\small}
\newtcolorbox{resultbox}[1][]{enhanced, unbreakable, colback=white, colframe=teal!60!black, boxrule=0.6pt, arc=2pt, left=5pt, right=5pt, top=3pt, bottom=3pt, fonttitle=\small\bfseries, colbacktitle=teal!10, coltitle=black, title={#1}, fontupper=\small}

\title{CARA: Causally Guided Retrieval Agent for Question Answering}

\author{
  \textbf{Weicheng Liu},
  \textbf{Xiangcheng Wu},
  \textbf{Depeng Xu},
  \textbf{Haopeng Zhang},
  \textbf{Xi Niu}
\\
  University of North Carolina at Charlotte
\\
  \small{\texttt{\{wliu23, xwu20, dxu7, hzhang55, xniu2\}@charlotte.edu}}
}

\begin{document}
\maketitle
\begin{abstract}
Retrieval-augmented generation over text-rich knowledge graphs requires identifying evidence that contributes to answering a question, rather than merely matching its textual content. Existing agentic retrievers adaptively balance global search and multi-hop exploration but primarily rely on textual similarity to guide retrieval. We propose CARA, a training-free, \textbf{CA}usally guided \textbf{R}etrieval \textbf{A}gent that explicitly evaluates the causal relevance of retrieved evidence through two complementary criteria: consequentiality, measuring whether adding evidence improves answerability, and essentiality, assessing whether removing evidence weakens answerability. CARA integrates these assessments with path consistency to guide graph exploration and employs reciprocal rank fusion to aggregate evidence from multiple agents. We evaluate CARA on three STaRK datasets and augment STaRK with ground-truth answers to enable downstream question answering evaluation. Experimental results demonstrate that CARA consistently outperforms existing retrievers. CARA also improves answer quality and faithfulness while examining fewer graph nodes and requiring fewer retrieval operations. These findings demonstrate that causal relevance assessment enables more focused and effective agentic graph retrieval, improving both retrieval performance and downstream question answering.

\end{abstract}

\section{Introduction}
As a promising direction in retrieval-augmented generation (RAG), retrieval over text-rich knowledge graphs (KGs) rather than flat documents is increasingly used to help large language models (LLMs) answer questions and ground answers in evidence. Unlike flat document retrieval, KG retrieval must jointly exploit nodes (breadth-oriented) and traverse along relations (depth-oriented). Balancing between the search breadth and depth is a long standing challenge for KG retrieval \citep{polonuer2026autonomous}. With the advances of agents, several recent studies \citep{sun2024think,yu2025graphflow} delegate such tasks to retrieval agents which have autonomy to balance search breadth and depth without a pre-specified number of node searches or hops.
However, these agents' retrieval process is driven primarily by textual similarity and graph structure, which can still favor evidence that is only associated with the query but not necessary for answering it. For example, consider the question, ``Which medicine should this patient avoid?'' A similarity-based retriever may return a clinical note stating, ``The patient currently takes several medicines, including aspirin and metformin,'' because it contains many words similar to the query. However, this information does not answer the question. Another record may simply state, ``Allergy: amoxicillin.'' Although this short record has no textual overlap with the question, it provides the key evidence needed to answer the question. This illustrates why textual similarity alone is not enough to make a substantive contribution to the answer.

In this study, we advocate for a causal retrieval agent, which has autonomy, but more importantly, is able to retrieve evidence that is ``causally relevant'' to answering the query, rather than merely textually similar. In this study, we operationalize the causal relevance using two complementary criteria:

\begin{itemize}
    \item \textit{Consequential}: does adding the retrieved evidence materially increase answerability?
    \item \textit{Essential}: does removing the retrieved evidence materially reduce answerability?
\end{itemize}

This operationalization is inspired by recent work on faithful rationale evaluation \citep{deyoung2020benchmark,yu2019rethinking}. It worthwhile to note that this causal relevance does not focus on the literal ``cause-effect'' relations, which is only one special case of this broader notion. 

In this paper, we design and implement a \textbf{CA}usally guided \textbf{R}etrieval \textbf{A}gent for question answering, named \textbf{CARA}, by retrieving causally relevant evidence, defined as consequential and essential evidence that is directly helpful to answer a query. After global anchoring and during neighborhood hop, the retrieval agent invokes a causal reasoning module, asking whether the current hop is consequential and essential for resolving the information need. This reasoning guides the agent's next decision on whether to continue hopping, to re-anchor globally, or to finish when sufficient support has been found.

Our contributions are threefold: (1) we formulate causal relevance as being consequential and essential, distinguishing answer-bearing evidence from merely textually similar evidence; (2) we introduce a training-free prompt-based causal reasoning module that evaluates the causal relevance from the perspectives of both the current hop and the accumulated path developed so far; and (3) we augment the well-known STaRK benchmark dataset with ground truth answers to the queries (questions), expanding its utility from only evaluating retrieval performance to evaluating both retrieval and question answering performance.

\section{Related Work}
Recent work on retrieval-augmented generation has increasingly moved beyond flat passage retrieval toward structured knowledge, multi-hop reasoning, and autonomous search. Three lines of research are particularly relevant to our work: knowledge-graph retrieval for question answering, causal retrieval and reasoning, and agentic retrieval.

\subsection{Knowledge Graph Retrieval for LLM Question Answering} 
Knowledge-graph retrieval represents entities and their relations explicitly, enabling LLMs to access evidence that is distributed across multiple connected pieces of information. Representative works include GraphRAG \citep{edge2024local}, KG2RAG \citep{zhu2025knowledge}, GRAG \citep{hu2025grag}, G-Retriever \citep{he2024g}, SubgraphRAG \citep{li2025simple}, KAR \citep{xia2025knowledge}, GraphSearch \citep{yang2025graphsearch}, HybGRAG \citep{lee2025hybgrag}, CoRAG \citep{zheng2025corag}, and MoR \citep{lei2025mixture}.

These approaches demonstrate the value of combining search with graph structure, but they primarily determine relevance through textual, relational, or structural compatibility. They do not explicitly ask whether a retrieved piece of evidence is consequential or essential to answering the query, which motivates our causal relevance formulation.

\subsection{Causal Reasoning for LLM Question Answering}
A growing body of work incorporates causal relations into retrieval and reasoning. Representative works include CDF-RAG \citep{khatibi2025cdf}, HCG-RAG \citep{saouda2026structure}, CausalAgent \citep{wang2026causalagent}, and CausalRAG \citep{wang2025causalrag}.

Although these approaches differ in implementation, they generally associate causality with cause-effect relations, dependency structures, or causal mechanisms represented in the underlying knowledge. Our notion is different. CARA does not require a retrieved edge or node to represent a literal cause-effect relation. Instead, we define causal relevance at retrieval time in terms of whether evidence is consequential and essential for answering the query. Thus, our focus is not causal structure in the world being described, but the causal contribution of retrieved evidence to the answering process.

\begin{figure*}
  \centering
  \includegraphics[width=\linewidth]{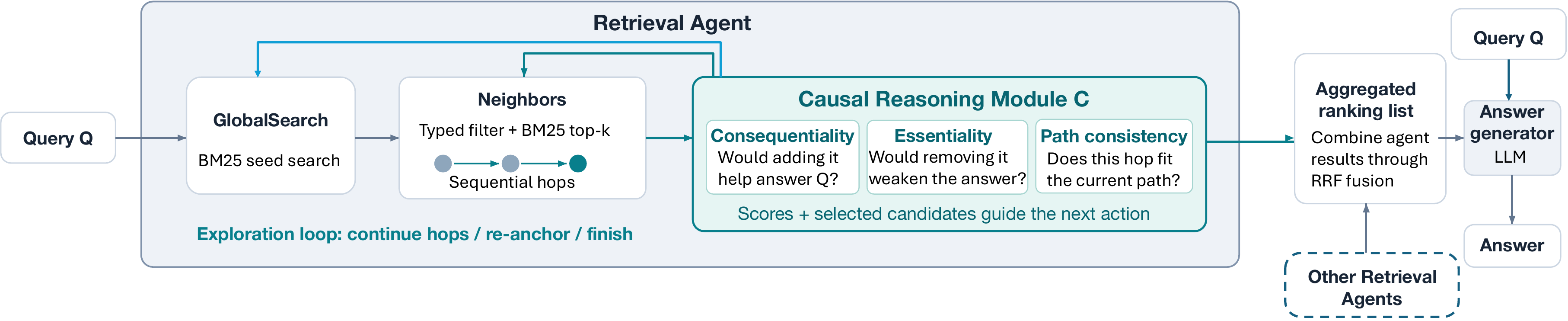}
  \caption{Overview of CARA. Each retrieval agent explores the knowledge graph with \textit{GlobalSearch} and \textit{Neighbors}. The causal reasoning module $\mathcal{C}$ assesses the neighborhood candidates. The scores guide the agent's next action. The agents' result lists are aggregated into one ranking, which the answer generator uses to answer $Q$.}
  \label{fig:overview}
\end{figure*}

\subsection{Agentic Retrieval for LLM Question Answering}
Agentic retrieval turns retrieval from a fixed pipeline into an adaptive decision process. Represented works include ReAct \citep{yao2023react}, AvaTaR \citep{wu2024avatar}, MA-RAG \citep{kumar2025multi}, SEMA-RAG \citep{huang2026sema}, and OmniSearch \citep{li2025benchmarking}.
Several recent methods extend agentic retrieval to knowledge graphs, including Tree-of-Traversals \citep{markowitz2024tree}, Think-on-Graph \citep{sun2024think}, Think-on-Graph 2.0 \citep{ma2025think}, RoG \citep{luo2024reasoning}, Search-on-Graph \citep{sun2026search}, and Explore-on-Graph \citep{yan2026explore}. Together, these methods establish graph retrieval as an adaptive search process in which an agent can decide where and how far to traverse.

However, existing agentic retrieval approaches are primarily based on textual similarity instead of causal retrieval we advocate for. Our work addresses this gap by introducing causal retrieval to agents in the broader sense of identifying evidence whose presence and absence materially contributes to resolving the information need.

\section{Method: CARA over Knowledge Graph}
The overview of CARA (\textbf{CA}usally guided \textbf{R}etrieval \textbf{A}gent) is in Figure \ref{fig:overview}. We study agentic retrieval over a knowledge graph $G=\langle V,E,\phi_V,\phi_E,d_V\rangle$, where $V$ and $E$ denote entities and edges, $\phi_V$ and $\phi_E$ assign a type to each node and relation, and $d_V(v)$ denotes the text attributes associated with node $v$, such as titles, descriptions, or other metadata. Given a natural language query (or question) $Q$, retrieval is formulated as a process in which an agent $A=\langle \mathrm{LLM},\mathcal{T},\mathcal{C}\rangle$ interacts with the graph through two retrieval tools \(\mathcal T\) and a causal reasoning module \(\mathcal C\). The retrieval tools include: 1) \textit{GlobalSearch} that conducts global node search and 2) \textit{Neighbors} that conducts one-hop neighborhood exploration. The causal module $\mathcal C$ is neither an independent agent nor a tool. It is an LLM-based reasoning and assessment module embedded within an agent. It analyzes the candidate nodes returned by the \textit{Neighbors} tool and provides causal assessments that inform the agent's subsequent decisions and actions.

The agent produces an interaction trajectory $ \tau= \big((s_1,A_1,o_1),\ldots,(s_T,A_T,o_T)\big), $ where at step \(t\), \(s_t\) contains 1) the original query \(Q\), 2) the agent interaction history \(I_t\), and 3) the current paths \(P_t\) that records one or multiple traversed paths so far. \(A_t\) denotes the agent's actions such as calling a retrieval tool, terminate, etc.  \(o_t\) is the output (a set of retrieved nodes $R_t$ and the corresponding paths $P_{t+1}$ leading to those nodes). At each step, the agent may \textit{Append} the newly retrieved nodes to \(R_t\), perform additional global or neighborhood retrieval, or terminate by issuing \textit{Finish}. Execution ends when the agent calls \textit{Finish} or when the maximum trajectory length \(T_{\max}\) is reached. 

\subsection{Retrieval Tools}

For an agent, there are two retrieval tools: 1) \textit{GlobalSearch} for seed nodes search (breadth-oriented search) and 2) \textit{Neighbors} for one-step hopping along the relations (depth-oriented search). 

\textit{GlobalSearch} retrieves the \(k\) highest-scoring nodes in the entire graph for an agent-generated subquery \(q\) based on the original $Q$:

\begin{equation}
    \mathcal G(q,k) = \operatorname{Top\text{-}k}_{v\in V} rel(q,d_V(v)).
\end{equation}

We implement \(rel()\) using BM25 \citep{robertson2009probabilistic} over an inverted index of node text attributes $d_V(v)$. The agent-generated subquery $q$ can evolve throughout exploration, allowing the agent to issue focused retrieval requests based on the trajectory state $s_t$. Global search is a breadth oriented search over KG, providing broad entry points into the graph or re-anchoring points when a local path becomes unproductive.

\textit{Neighbors} retrieves adjacent (one-hop neighbor) nodes subject to optional node- and edge-type constraints
$F_t=(F_{V,t},F_{E,t})$ selected by the agent. The result is a filtered set $ N_{F_t}(v_t)$ of one-hop neighborhood of the current node $v_t$. To further trim this filtered set, this \textit{Neighbors} tool conducts another BM25 search to keep the top $k$ from this filtered set:

\begin{equation}
\begin{aligned}
    \mathcal N_t
    &= \operatorname{Neighbors}(v_t,q_t,F_t) \\
    &= \operatorname{Top\text{-}k}_{u\in N_{F_t}(v_t)}
       rel(q_t,d_V(u)).
\end{aligned}
\end{equation}

Here $ \mathcal N_t=\{u_1,\ldots,u_k\}$ denotes the candidate neighbors retrieved at hop \(t\). The repeated applications of the \textit{Neighbors} tool compose into multi-hop traversal. \textit{GlobalSearch} remains available along the way for re-anchoring. Unlike traditional non-agentic retrieval, the number of hops is not specified in advance. Instead, the agent decides whether to continue hopping, re-anchor globally, or finish based on the accumulated evidence. It is noteworthy that each candidate neighborhood node $u_i$, will further be passed through the causal reasoning module $\mathcal{C}$ before the agent chooses its next action. 

\subsection{Causal Reasoning Module}

\textit{Neighbors} retrieval identifies nodes that are structurally adjacent and textually similar to the current exploration state. However, neither graph adjacency nor textual similarity guarantees that a candidate node provides evidence that materially contributes to answering the query. Therefore we introduce a causal reasoning module $\mathcal C$ that evaluates each candidate according to its consequentiality and essentiality for resolving the information need.

Given the candidate set \(\mathcal N_t=\{u_1,\ldots,u_k\}\) returned by the \textit{Neighbors} tool, the module receives the query \(Q\), the current node \(v_t\), the candidate nodes \(\mathcal N_t\), their connecting relations, and the accumulated paths \(P_{t+1}\). Unlike literal causal reasoning that seeks cause-effect relations in the underlying domain, our notion of causal relevance is retrieval-oriented: evidence is considered causally relevant when its presence or absence materially affects the ability to support an answer to \(Q\). This operationalization is motivated by two complementary properties: consequentiality and essentiality.

For each candidate \(u_i\), the module $\mathcal C$ estimates a consequentiality score \(c_i\in[0,1]\) and an essentiality score \(e_i\in[0,1]\). Consequentiality asks whether incorporating the candidate node and its associated hop materially changes the evidence available for answering the query. It captures the candidate's incremental answer contribution, rather than simply its textual similarity. 

The LLM is therefore asked whether adding \(u_i\), together with its relation to \(v_t\), introduces information that substantially strengthens or advances the answer. 

Essentiality instead asks whether removing the candidate evidence from the currently available evidence would materially reduce the ability to justify an answer. 

More directly, the LLM is prompted to reason counterfactually: if this candidate node and its connecting relation were unavailable, would an important part of the evidence needed to answer the query be lost? A high \(e_i\) therefore indicates that the candidate carries information that is difficult to replace with evidence already contained in the current trajectory.

We combine these two judgments into a node-level causal relevance score:
\begin{equation}
    cau_i=\lambda c_i+(1-\lambda)e_i,
\end{equation}
where \(\lambda\in[0,1]\) controls the relative emphasis on consequentiality and essentiality. Importantly, \(cau_i\) should not be interpreted as a probability that a literal causal relationship exists between \(v_t\) and \(u_i\). Rather, it measures whether the candidate evidence has a materially incremental and counterfactual contribution to answering \(Q\).

Causal relevance alone, however, evaluates a candidate primarily with respect to the query and the evidence currently available. In multi-hop retrieval, a locally useful candidate may still lead the agent away from the reasoning trajectory developed so far. We therefore introduce a separate path consistency score \(p_i\in[0,1]\). Unlike \(cau_i\), which measures the node-level contribution to answering the query, \(p_i\) evaluates whether the transition \(v_t\rightarrow u_i\) forms a coherent continuation of the accumulated path developed so far. In particular, the module examines whether the relation connecting \(v_t\) and \(u_i\) is compatible with the information need represented by the current trajectory and whether following this candidate preserves a plausible chain of evidence toward the answer.

The final score for candidate node \(u_i\) is:
\begin{equation}
    s_t(u_i)=\alpha cau_i+(1-\alpha) p_i,
\end{equation}
where \(\alpha \in[0,1]\) controls the relative importance of node-level causal relevance and the trajectory-level consistency. The highest-scoring candidates are returned to the retrieval agent.

Importantly, the causal reasoning module $\mathcal{C}$ is an assessment module rather than an autonomous agent. It does not independently invoke \textit{GlobalSearch} or \textit{Neighbors}. Instead, its assessments enhance the information presented to the retrieval agent and therefore influence subsequent search decisions. Based on the causal and path-consistency assessments, the agent may continue along a promising neighborhood path with \textit{Neighbors}, reformulate the subquery, modify node or relation filters, re-anchor through \textit{GlobalSearch}, \textit{Append} nodes to the retrieved set, or terminate with \textit{Finish} when additional exploration is unlikely to provide consequential evidence. In this way, causal reasoning affects not merely the reranking of already retrieved nodes, but the direction and termination of the retrieval process itself.

\subsection{Aggregation of Each Agent Result}

To improve robustness and reduce dependence on a single retrieval trajectory, we run multiple agents in parallel. Each agent \(j\) independently interacts with the knowledge graph, producing a retrieval trajectory \(\tau^{(j)}\) and an ordered list of retrieved nodes $R^{(j)}$. Independent trajectories may differ in their initial anchors, generated subqueries, neighborhood expansions, and traversal decisions. Such diversity allows different agents to explore alternative paths through the graph and reduces sensitivity to any single search trajectory. 

Rather than directly concatenating the node lists from different agents, inspired by the classic idea of reciprocal rank fusion (RRF) \citep{cormack2009reciprocal}, we aggregate them in the following way. For a candidate node \(u\), its cross-agent fusion score is:
\begin{equation}
    \mathrm{RRF}(u) = \sum_{j:u\in R^{(j)}} \frac{1}{\mathrm{rank}_j(u)},
\end{equation}
where \(\mathrm{rank}_j(u)\) denotes the position of node \(u\) in the ranked evidence list $R^{(j)}$ returned by agent \(j\). In this new context, RRF captures two complementary properties of the retrieved evidence. First, it rewards high within-agent rank. Second, it rewards cross-agent consensus. Thus, evidence that is repeatedly identified across different trajectories receives stronger aggregate support than evidence found by only one trajectory. The final candidate set is formed from the union of the nodes retrieved by all agents,

ranked in descending order of their RRF scores. 
In the rare case that two nodes receive identical RRF scores, the node with the better best rank across agents is placed first.

\section{Experiment Setup}

We evaluate CARA from two complementary perspectives: retrieval performance and question answering performance.

\subsection{STaRK and Ground-Truth Answer Augmentation}
We evaluate on the three text-rich knowledge graphs in the well-known benchmark STaRK \citep{wu2024stark}. It contains three datasets: AMAZON, MAG, and PRIME. AMAZON represents an e-commerce domain, MAG a scholarly domain, and PRIME a biomedical domain. STaRK provides natural-language queries $Q$ together with ground-truth entities relevant to answer $Q$, allowing retrieval systems to be evaluated against known nodes. However, it does not provide a ground truth for evaluating generated answers. We therefore augment STaRK with ground-truth answers. For each query $Q$, we use the ground truth entities' text attributes $d_V(v)$ and related relations as the source information and employ Qwen3.8-27B to generate answers. We instruct the model in the prompt to satisfy four requirements: \textit{Relevance}, ensuring that the answer directly addresses $Q$; \textit{Faithfulness}, ensuring that all factual claims are supported by the provided entity information; \textit{Coverage}, ensuring that the names of all ground-truth entities are explicitly mentioned; and \textit{Naturalness}, ensuring that the answer is fluent and natural rather than enumerating a list of retrieved entities. 
After answer generation, we took two extra steps to validate the answers. First, we validate the answer using an automatically calculated indicator \textit{ Completeness} defined as the proportion of ground-truth entities whose complete names are explicitly mentioned in the generated answer. If \textit{Completeness} is below a threshold, we prompt the model to regenerate a new version of answer until it reaches the threshold. We repeat this generation-and-validation process, resulting in more complete ground truth answers. Second, we manually inspect stratified samples (approximately 10\% of the generated answers) across AMAZON, MAG, and PRIME to validate that the resulting ground truth answers are appropriate across different domains and query types.

\subsection{Baselines}

We compare CARA with conventional retrievers and agentic retrievers used in prior work. The conventional retrievers include 
the classic sparse retrieval, BM25 \citep{robertson2009probabilistic}; dense embedding retrieval that rank nodes by cosine similarity, ada-002 \citep{openai2022ada} and GritLM-7B \citep{muennighoff2025gritlm}; knowledge-aware retrieval KAR \citep{xia2025knowledge} that conducts query expansion and applies relation type constraints during retrieval. Agentic retrievers include Think-on-Graph \citep{sun2024think}, which is a training-free LLM agent that iteratively expands paths in the graph using beam search; GraphFlow \citep{yu2025graphflow}, which learns to generate multi-hop retrieval trajectories using GFlowNets \citep{bengio2021flow}; and ARK \citep{polonuer2026autonomous}, which has adaptive use of global search and neighborhood exploration but without causal reasoning during graph traversal. 

All agentic systems (CARA and agentic baselines), the causal module, the common answer generator use the same open-weight backbone, Qwen3.8-27B, served with vLLM with thinking disabled. The agents' LLM calls sample with temperature 1.0, and all other calls decode greedily. CARA runs $m=3$ agents per query (question) with $T_{\max}=150$ steps each; \textit{GlobalSearch} returns $k=30$ candidates and \textit{Neighbors} returns $k=20$.

\subsection{Retrieval Evaluation}
We first evaluate the performance of the entity-retrieval task. We report Hit@1, Recall@20, and Mean Reciprocal Rank (MRR). These metrics capture complementary aspects of retrieval performance: the proportion of a correct entity ranked first, the proportion of ground truth entities successfully retrieved among the top 20 ranked results, and how highly the first correct entity is ranked overall.

\begin{table*}[t]
\centering
\small
\setlength{\tabcolsep}{5.5pt}
\renewcommand{\arraystretch}{1.12}
\begin{tabular}{@{}l@{\hspace{10pt}}ccc@{\hspace{12pt}}ccc@{\hspace{12pt}}ccc@{}}
\toprule
 & \multicolumn{3}{c}{\textsc{Amazon}} & \multicolumn{3}{c}{\textsc{MAG}} & \multicolumn{3}{c}{\textsc{Prime}} \\
\cmidrule(lr){2-4} \cmidrule(lr){5-7} \cmidrule(l){8-10}
\textbf{Method} & Hit@1 & R@20 & MRR & Hit@1 & R@20 & MRR & Hit@1 & R@20 & MRR \\
\midrule
\multicolumn{10}{@{}l}{\textit{Conventional retrievers}} \\
\quad BM25 & 45.3 & 53.7 & 55.4 & 25.9 & 45.7 & 34.9 & 12.7 & 31.3 & 19.8 \\
\quad ada-002 & 39.7 & 53.8 & 50.9 & 29.1 & 48.4 & 38.6 & 12.6 & 36.0 & 21.3 \\
\quad GritLM-7B & 49.8 & \textbf{61.8} & 60.5 & 39.1 & 61.4 & 50.5 & 15.0 & 38.4 & 23.4 \\
\quad KAR & 49.6 & 59.7 & 59.0 & 50.1 & 63.9 & 58.6 & 23.3 & 46.1 & 32.2 \\
\midrule
\multicolumn{10}{@{}l}{\textit{Agentic retrievers}} \\
\quad Think-on-Graph & 22.2 & 27.8 & 29.2 & 36.8 & 52.7 & 43.6 & 15.8 & 37.8 & 23.5 \\
\quad Think-on-Graph 2.0 & 48.8 & 37.3 & 54.6 & 43.9 & 50.8 & 48.6 & 23.8 & 43.8 & 30.9 \\
\quad GraphFlow$^\dagger$ & 47.9 & 36.2 & 55.5 & 39.1 & 57.2 & 47.8 & 51.4 & \textbf{79.7} & 61.4 \\
\quad ARK & \underline{59.0} & 60.2 & \underline{67.7} & \underline{80.1} & \underline{77.9} & \underline{83.1} & \underline{58.6} & 71.9 & \underline{65.3} \\
\rowcolor{gray!12} \quad \textbf{CARA (ours)$^{*}$} & \textbf{62.4} & \underline{60.5} & \textbf{71.3} & \textbf{84.9} & \textbf{80.8} & \textbf{88.1} & \textbf{59.5} & \underline{73.4} & \textbf{67.7} \\
\bottomrule
\end{tabular}
\caption{Retrieval performance (\%) on the full STaRK test sets (\textsc{Amazon} $n{=}1{,}642$, \textsc{MAG} $n{=}2{,}665$, \textsc{Prime} $n{=}2{,}801$). $^\dagger$Reported by \citet{yu2025graphflow} (LLaMA3-8B-Instruct, test-0.1 subset). $^{*}$CARA version: $\lambda = 0.5, \alpha = 0.5$. \textbf{Bold}: best; \underline{underline}: second best.}
\label{tab:stark-main-results}
\end{table*}

\subsection{Question Answering Evaluation}
We next evaluate whether the retrieved evidence improves downstream question answering. To isolate the effect of retrieval, all retrieval systems use the same answer generator (LLM, Qwen3.8-27B), prompt, decoding configuration, and context budget. The only difference between systems is the retrieved evidence supplied to the answer generator. In this end-to-end setting, each retriever produces its own ranked evidence set, and the common generator answers the question using that evidence. To test whether better retrieval leads to better answers, we evaluate answer correctness using BERTScore \citep{zhang2020bertscore} between the generated answer and the ground truth answer. We also evaluate answer faithfulness using the Ragas framework \citep{es2024ragas}, measuring whether factual claims in the generated answer are supported by the retrieved context. %

\begin{table}[t]
\centering
\small
\setlength{\tabcolsep}{2.2pt}
\renewcommand{\arraystretch}{1.12}
\begin{tabular}{@{}l@{\hspace{3pt}}cc@{\hspace{4pt}}cc@{\hspace{4pt}}cc@{}}
\toprule
 & \multicolumn{2}{c}{\textsc{Amazon}} & \multicolumn{2}{c}{\textsc{MAG}} & \multicolumn{2}{c}{\textsc{Prime}} \\
\cmidrule(lr){2-3} \cmidrule(lr){4-5} \cmidrule(l){6-7}
\textbf{Method} & BS & Faith. & BS & Faith. & BS & Faith. \\
\midrule
No retrieval & 7.5 & -- & 16.7 & -- & 27.9 & -- \\
KAR & 28.3 & \textbf{88.3} & 33.0 & \textbf{86.6} & 36.0 & \textbf{88.6} \\
ARK & \underline{30.1} & 87.4 & \underline{39.2} & 78.6 & \underline{40.5} & 80.5 \\
\rowcolor{gray!12} \textbf{CARA (ours)$^{*}$} & \textbf{30.7} & \underline{88.0} & \textbf{40.2} & \underline{82.6} & \textbf{41.3} & \underline{86.1} \\
\bottomrule
\end{tabular}
\caption{Question answering on the full STaRK test sets. All methods share the same answer generator. BERTScore (BS, $\times$100) compares each answer with the ground-truth answer. Faithfulness (Faith., \%) is the fraction of answer claims supported by the retrieved context. Faithfulness is not defined in the ``No retrieval'' approach. $^{*}$CARA version: $\lambda = 0.5, \alpha = 0.5$. \textbf{Bold}: best; \underline{underline}: second best.}
\label{tab:stark-qa-results}
\end{table}

\section{Results}

\subsection{Retrieval Performance}
As the first step of experiment, we tried different $\lambda$ values from 0 to 1 (in Equation 3, balancing consequentiality and essentiality) and $\alpha$ values from 0 to 1 (in Equation 4, balancing node-level causal relevance and path consistency). It is worthnoting that the setting of $\lambda = 0\ or\ 1$ or $\alpha = 0\ or\ 1$ is equivalent to a component-level ablation study. Averaged across datasets and metrics, $\lambda = 0.5$ and $\alpha = 0.5$ achieve the highest performance, supporting the selection of these values.

Table~\ref{tab:stark-main-results} compares CARA ($\lambda = 0.5$, $\alpha = 0.5$) with conventional and agentic retrievers on the three STaRK datasets. CARA achieves the highest Hit@1 and MRR across AMAZON, MAG, and PRIME, demonstrating its effectiveness in ranking relevant entities. Compared with ARK, its closest agentic baseline, CARA's improvements are particularly pronounced on MAG, whose questions tend to be relation-heavy. These findings suggest that incorporating causal relevance assessment and the rank-based aggregation into adaptive graph exploration improves retrieval effectiveness across diverse knowledge domains, especially in relation-heavy domains.

\subsection{Question Answering Performance}
Table~\ref{tab:stark-qa-results} evaluates the downstream question answering performance of CARA against KAR and ARK using BERTScore and the RAGAS faithfulness. CARA achieves the highest BERTScore across all three datasets, demonstrating that improved retrieval effectiveness translates into better answer quality. CARA also consistently improves faithfulness over ARK, with particularly notable gains on MAG and PRIME. Interestingly, KAR achieves the highest faithfulness across all datasets despite substantially lower BERTScore values. This suggests that generating answers closely grounded in retrieved evidence does not necessarily guarantee their correctness or completeness relative to ground-truth answers. Overall, these findings highlight the importance of evaluating both answer quality and faithfulness, and demonstrate that causal-guided retrieval improves downstream question answering while maintaining strong evidence grounding.

\begin{figure}[t]
  \centering
  \includegraphics[width=\columnwidth]{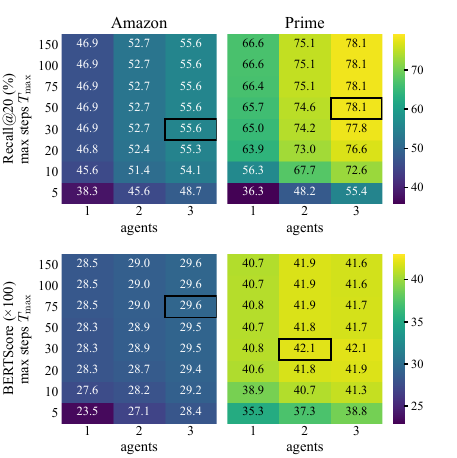}
  \caption{Retrieval (Recall@20, \%) and answer quality (BERTScore, $\times$100) of CARA on \textsc{Amazon} and \textsc{Prime} under different exploration budgets: maximum steps per agent $T_{\max}$ (rows) and number of agents (columns). The outlined cell marks each panel's peak (earliest budget when tied). All values are on the STaRK test-0.1 subset (10\% of each test set).}
  \label{fig:budget}
\end{figure}
\subsection{Impact of Exploration Budget on Performance}
Figure~\ref{fig:budget} investigates how CARA's retrieval and question answering performance varies with the exploration budget, controlled by the number of agents (1–3) and maximum exploration steps ($T_{\max}$), ranging from 5 to 150. Overall, both Recall@20 and BERTScore improve substantially as the exploration budget increases, but exhibit diminishing returns beyond a certain point. Increasing the number of agents consistently improves Recall@20 for both AMAZON and PRIME datasets, highlighting the benefits of exploring diverse retrieval trajectories. However, further increasing $T_{\max}$ to 150 produces only marginal improvements, suggesting that most relevant evidence is discovered relatively early during exploration.

Interestingly, question answering performance exhibits a similar but less consistent trend. With three agents, BERTScore improves rapidly at smaller exploration budgets and largely stabilizes beyond $T_{\max}=30$. On PRIME, BERTScore even decreases slightly from 42.1 at $T_{\max}=30$ to 41.6 at $T_{\max}=150$, suggesting that retrieving additional evidence does not necessarily improve answer quality. %

\subsection{Ablation Study}
\label{sec:ablation}
Table~\ref{tab:retrieval-qa-results} investigates the contributions of the causal reasoning module ($\mathcal{C}$) and reciprocal rank fusion (RRF), compared with the count-based aggregation used by ARK. Removing $\mathcal{C}$ consistently degrades retrieval and question answering performance. Replacing RRF with ARK's count-based aggregation yields smaller performance differences. However, count-based aggregation achieves marginally higher Hit@1 on PRIME and faithfulness on MAG and PRIME. Overall, these results demonstrate that causal reasoning provides the primary performance gains, while RRF offers additional improvements in evidence ranking and answer quality.

\begin{table*}[t]
\centering
\footnotesize
\setlength{\tabcolsep}{2.2pt}
\renewcommand{\arraystretch}{1.12}
\begin{tabular}{@{}l@{\hspace{6pt}}ccccc@{\hspace{6pt}}ccccc@{\hspace{6pt}}ccccc@{}}
\toprule
 & \multicolumn{5}{c}{\textsc{Amazon} ($n{=}1{,}642$)} & \multicolumn{5}{c}{\textsc{MAG} ($n{=}2{,}665$)} & \multicolumn{5}{c}{\textsc{Prime} ($n{=}2{,}801$)} \\
\cmidrule(lr){2-6} \cmidrule(lr){7-11} \cmidrule(l){12-16}
 & \multicolumn{3}{c}{Retrieval} & \multicolumn{2}{c}{QA} & \multicolumn{3}{c}{Retrieval} & \multicolumn{2}{c}{QA} & \multicolumn{3}{c}{Retrieval} & \multicolumn{2}{c}{QA} \\
\cmidrule(lr){2-4} \cmidrule(lr){5-6} \cmidrule(lr){7-9} \cmidrule(lr){10-11} \cmidrule(lr){12-14} \cmidrule(l){15-16}
\textbf{Variant} & Hit@1 & R@20 & MRR & BS & Faith. & Hit@1 & R@20 & MRR & BS & Faith. & Hit@1 & R@20 & MRR & BS & Faith. \\
\midrule
\rowcolor{gray!12} \textbf{CARA (ours)$^{*}$} & \textbf{62.4} & \textbf{60.5} & \textbf{71.3} & \textbf{30.7} & \textbf{88.0} & \textbf{84.9} & \textbf{80.8} & \textbf{88.1} & \textbf{40.2} & \underline{82.6} & \underline{59.5} & \textbf{73.4} & \textbf{67.7} & \textbf{41.3} & \underline{86.1} \\
w/o $\mathcal{C}$ & 60.5 & \underline{60.2} & 69.1 & 30.4 & 87.7 & 80.3 & \underline{77.9} & 83.5 & 39.2 & 78.5 & 58.1 & 72.1 & 65.4 & 40.8 & 80.7 \\
count-based & \underline{62.2} & \textbf{60.5} & \underline{70.3} & \underline{30.6} & \textbf{88.0} & \underline{84.4} & \textbf{80.8} & \underline{87.4} & \underline{40.1} & \textbf{82.8} & \textbf{60.0} & \underline{73.1} & \underline{67.2} & \underline{40.9} & \textbf{86.3} \\
w/o $\mathcal{C}$, count-based & 59.0 & \underline{60.2} & 67.7 & 30.1 & 87.4 & 80.1 & \underline{77.9} & 83.1 & 39.2 & 78.6 & 58.6 & 71.9 & 65.3 & 40.5 & 80.5 \\
\bottomrule
\end{tabular}
\caption{Ablation of the causal reasoning module $\mathcal{C}$ and the rank-based aggregation (RRF) on the full STaRK test sets. w/o $\mathcal{C}$: agents without $\mathcal{C}$. count-based: nodes ranked by the frequency of agents' votes instead of RRF; w/o $\mathcal{C}$, count-based is equivalent to ARK (Tables~\ref{tab:stark-main-results} and~\ref{tab:stark-qa-results}). $^{*}$CARA version: $\lambda = 0.5, \alpha = 0.5$. \textbf{Bold}: best; \underline{underline}: second best.}
\label{tab:retrieval-qa-results}
\end{table*} 
\begin{table*}
\centering
\small
\setlength{\tabcolsep}{5pt}
\renewcommand{\arraystretch}{1.12}
\begin{tabular}{@{}ll@{\hspace{10pt}}ccc@{\hspace{12pt}}c@{\hspace{12pt}}cc@{\hspace{12pt}}c@{\hspace{12pt}}c@{}}
\toprule
 & & \multicolumn{3}{c}{\textbf{LLM calls}} & \textbf{Tokens} & \multicolumn{2}{c}{\textbf{Tool calls}} & \textbf{Nodes} & \textbf{Latency} \\
\cmidrule(lr){3-5} \cmidrule(lr){7-8}
\textbf{Dataset} & \textbf{Method} & Agent & $\mathcal{C}$ module & Total & (k) & GlobalSearch & Neighbors & \textbf{examined} & (s) \\
\midrule
\textsc{Amazon} & ARK & 26.9 & -- & \textbf{26.9} & 310 & 18.2 & 15.2 & 152 & \textbf{105} \\
\rowcolor{gray!12} \cellcolor{white} & \textbf{CARA (ours)$^{*}$} & \textbf{22.1} & 8.4 & 30.6 & \textbf{244} & \textbf{13.2} & \textbf{10.9} & \textbf{122} & 145 \\
\midrule
\textsc{MAG} & ARK & 31.5 & -- & \textbf{31.5} & \textbf{338} & 16.1 & 30.4 & 163 & \textbf{136} \\
\rowcolor{gray!12} \cellcolor{white} & \textbf{CARA (ours)$^{*}$} & \textbf{25.8} & 20.1 & 45.9 & 340 & \textbf{11.9} & \textbf{22.3} & \textbf{123} & 223 \\
\midrule
\textsc{Prime} & ARK & 41.2 & -- & \textbf{41.2} & \textbf{469} & 20.2 & 47.8 & 249 & \textbf{141} \\
\rowcolor{gray!12} \cellcolor{white} & \textbf{CARA (ours)$^{*}$} & \textbf{32.4} & 24.1 & 56.5 & 511 & \textbf{14.9} & \textbf{32.1} & \textbf{198} & 276 \\
\bottomrule
\end{tabular}
\caption{Per-query cost on the full STaRK test sets, summed over the three agents. Lower is better. \emph{Total} counts agent calls and $\mathcal{C}$ module calls. Tokens cover these calls. 
Latency is the mean wall-clock time per query on Nvidia H200 GPUs (Qwen3.8-27B served with vLLM).$^{*}$CARA version: $\lambda = 0.5, \alpha = 0.5$.}
\label{tab:efficiency-results}
\end{table*}

\subsection{Efficiency and Exploration Behavior}
\label{sec:efficiency}
Table~\ref{tab:efficiency-results} compares the computational efficiency and exploration behavior of CARA and ARK across three datasets. Interestingly, CARA achieves more focused graph exploration despite introducing an additional causal reasoning module. Compared with ARK (the closest agentic baseline), CARA reduces \textit{GlobalSearch} calls as well as \textit{Neighbors} calls across the datasets. The number of examined nodes also decreases for the three datasets. These results suggest that causal relevance assessment helps agents identify promising evidence paths while avoiding unnecessary exploration. 

Compared with AMAZON whose questions are more entity-oriented, MAG and PRIME involve more relation-intensive biomedical questions that require deeper multi-hop exploration. This difference is reflected in CARA's substantially higher number of \textit{Neighbors} calls and therefore causal reasoning calls per query, suggesting that causal guidance is particularly important for navigating complex relational structures. However, the additional causal reasoning introduces latency overhead. The particularly pronounced overhead on MAG and PRIME highlights a trade-off between retrieval effectiveness and computational efficiency, suggesting opportunities for reducing latency through parallel or batched causal assessments.

\section{Conclusion}
In this paper, we proposed CARA, a training-free causally guided retrieval agent for question answering over text-rich knowledge graphs. Unlike existing agentic retrievers that primarily rely on textual similarity, CARA evaluates retrieved evidence through consequentiality and essentiality, together with path consistency, to guide graph exploration. Reciprocal rank fusion further aggregates evidence from multiple independent agents. We also augmented the STaRK benchmark with ground-truth answers to enable downstream question answering evaluation. Experiments on three datasets demonstrate that CARA consistently improves retrieval effectiveness, while also improving answer quality and faithfulness compared with ARK. Moreover, CARA achieves more focused exploration by examining fewer nodes and requiring fewer retrieval operations. However, the additional causal reasoning introduces inference latency overhead, particularly for relation-intensive questions. 

\section*{Limitations}
Despite its promising results, CARA has several limitations. First, causal relevance is operationalized through LLM-based assessments of consequentiality and essentiality rather than formal inference. These assessments may be sensitive to model biases, prompting strategies, and errors in evaluating evidence contributions. Second, although CARA reduces retrieval operations and examined nodes, the additional causal reasoning module increases computational overhead and inference latency, particularly for relation-intensive questions. Third, downstream question answering evaluation relies on automatically constructed ground-truth answers and LLM-based faithfulness assessments, which may introduce evaluation noise. Future work will explore more efficient causal assessments and alternative evaluations.

\bibliography{custom}

\clearpage

\appendix

\section{Appendix}
\subsection{Implementation Details}
\label{app:impl}\label{sec:appendix}

Table~\ref{tab:impl-settings} lists the settings of the implementation. 

\begin{table}[t]
\centering
\footnotesize
\setlength{\tabcolsep}{3pt}
\renewcommand{\arraystretch}{1.08}
\begin{tabular}{@{}>{\raggedright\arraybackslash}p{0.27\columnwidth}>{\raggedright\arraybackslash}p{0.70\columnwidth}@{}}
\toprule
\rowcolor{gray!12}\multicolumn{2}{@{}l}{\textbf{Backbone}}\\
LLM & Qwen3.8-27B for every call, thinking mode off\\
Serving & vLLM 0.30.0; 6 replicas, 1 H200 each\\
\rowcolor{gray!12}\multicolumn{2}{@{}l}{\textbf{Agents}}\\
Agents & $m=3$ per question, run in parallel\\
Step budget & $T_{\max}=150$ LLM calls per agent\\
Decoding & temperature 1.0, top-$p$ 0.95, top-$k$ 20\\
Actions & \textit{GlobalSearch}, \textit{Neighbors}, \textit{Append}, \textit{Finish}\\
\rowcolor{gray!12}\multicolumn{2}{@{}l}{\textbf{Tools}}\\
\textit{GlobalSearch} & BM25 over node names and descriptions; up to $k=30$ results\\
\textit{Neighbors} & one hop; type filters; BM25 subquery; $k=20$ shown\\
\rowcolor{gray!12}\multicolumn{2}{@{}l}{\textbf{Causal module $\mathcal{C}$}}\\
Calls & one per \textit{Neighbors} call with candidates, all at once\\
Output & $c_i, e_i, p_i\in[0,1]$ per candidate\\
Decoding & temperature 0, at most 1,024 tokens\\
Weights & $\lambda=0.5$, $\alpha=0.5$\\
\rowcolor{gray!12}\multicolumn{2}{@{}l}{\textbf{Aggregation}}\\
Rule & RRF $\sum_j 1/\mathrm{rank}_j(u)$\\
Ties & best rank, then first position\\
\rowcolor{gray!12}\multicolumn{2}{@{}l}{\textbf{QA evaluation}}\\
Gen.\ input & top-5 nodes, evidence body $\le$1,500 characters each\\
Gen.\ decoding & temperature 0, at most 2,000 tokens\\
BERTScore & F1, roberta-large, rescaled, $\times$100\\
Faithfulness & RAGAS 0.4.3; judge Qwen3.8-27B; mean of 2 passes; empty context $=0$\\
Test questions & Amazon 1,642; MAG 2,665; Prime 2,801\\
QA questions & shared by all methods\\
\bottomrule
\end{tabular}
\caption{Settings of the implementation}
\label{tab:impl-settings}
\end{table}

\subsection{Prompts of the Causal Reasoning Module}
\label{app:prompts}

\begin{promptbox}[Prompt of the causal reasoning module $\mathcal{C}$]
You are the causal reasoning module $\mathcal{C}$ of a retrieval agent that explores a knowledge graph to answer a query. You are an assessment module, not an agent: you do not search the graph and you do not choose the agent's next action. The agent has just used the \textit{Neighbors} tool on the current node; assess each returned candidate node.

\textbf{What ``causal'' means here}

* Causal relevance is retrieval-oriented. Evidence is causally relevant when its presence or absence materially affects the ability to support an answer to the query. It is not a cause-effect relation in the domain.

\textbf{Task}

\textbf{You are given four inputs:}

\textbf{1. Query $Q$}

* The natural-language query the agent is answering.

\textbf{2. Accumulated paths $P_{t+1}$}

* The paths the agent has traversed so far, ending at the current node.

\textbf{3. Current node $v_t$}

* The node the agent has just expanded, with its type and text description.

\textbf{4. Candidate set $\mathcal{N}_t=\{u_1,\ldots,u_k\}$}

* The candidate neighbors returned by \textit{Neighbors}, each with its type, its connecting relations to $v_t$ and its text description.

\textbf{Assess every candidate $u_i$ with three scores in $[0,1]$.}

1. \textit{Consequentiality $c_i$: the candidate's incremental answer contribution.}

* Would adding $u_i$, together with its relation to $v_t$, introduce information that substantially strengthens or advances the answer to $Q$, beyond what the accumulated paths already provide?

* This is not textual similarity: a candidate that repeats known evidence, or is only on the same topic as $Q$, scores low even if it shares many words with $Q$.

2. \textit{Essentiality $e_i$: the counterfactual loss without the candidate.}

* If $u_i$ and its connecting relation were unavailable, would an important part of the evidence needed to answer $Q$ be lost?

* This is not the general importance of the node: score high only when the evidence is difficult to replace with evidence already contained in the current trajectory, and low when an equivalent piece is already on a path.

3. \textit{Path consistency $p_i$: the coherence of the step $v_t\rightarrow u_i$.}

* Does the transition form a coherent continuation of the accumulated paths? Is the connecting relation compatible with the information need in $Q$ (required entity types, relations and named entities), and does following $u_i$ preserve a plausible chain of evidence toward the answer?

* This is not mere adjacency: every candidate is already a neighbor of $v_t$.

\textbf{Scoring Scale} (the same for all three scores)

* 0.0--0.2: unrelated or fully redundant with the accumulated paths.

* 0.3--0.4: weak. \quad 0.5--0.6: moderate. \quad 0.7--0.8: strong.

* 0.9--1.0: decisive, e.g.\ the answer entity, an indispensable bridge, or evidence without which $Q$ cannot be answered.

\textbf{Rules}

1. \textit{Neither adjacency nor textual similarity is evidence by itself.}

* Candidates are listed in the order the \textit{Neighbors} tool returned them; this order carries no evidential weight.

2. \textit{Use only the given inputs.}

* Base every score on the query, the accumulated paths, the connecting relations and the candidate descriptions.

3. \textit{Assess each candidate independently.}

* The scores need not sum to anything. Do not predict the final answer.

4. \textit{Assess every candidate.}

* Return a line for every candidate, including irrelevant ones, which receive low scores.

\textbf{Output Format}

Return one line per candidate, in the given order:

node\_index \textbar{} $c_i$ \textbar{} $e_i$ \textbar{} $p_i$

Do not output anything else.

\end{promptbox}

\subsection{Prompts for Ground-Truth Answer Augmentation}
\label{app:gtanswer}

\begin{promptbox}[Prompt for AMAZON dataset]
You are a grounded natural-language answer generation model in a retrieval-and-generation system.

\textbf{Task}

\textbf{You are given two inputs:}

\textbf{1. User Query}

* A natural-language query representing the user's information need or search intent.

\textbf{2. Ground-Truth Products}

* A set of one or more ground-truth products associated with the user query.

* Each product may contain information such as: product name, brand, dimensions, description, feature, review, Q\&A, or other product-related information. If the information is ``nan'', it is unknown and do not use it.

* The ground-truth products are the ONLY source of product-related factual information that may be used when generating the answer.

\textbf{Generate an answer for the user query based on the provided information.}

* Answer the query directly and concisely in a natural and conversational way, rather than sounding like a bibliography, database entry, or search result.

* Use the complete product name exactly as provided in the input whenever a product is mentioned. Do not abbreviate, truncate, paraphrase, normalize, or replace the product name.

* Present the products naturally as part of a coherent answer rather than as a raw list. Do not use bullet points, numbered lists, tables, headings, or any other list-like format.

* Write the entire answer as one continuous paragraph without line breaks or separate sections.

* Use only factual information explicitly provided for the corresponding product.

\textbf{Grounding and Factuality Requirements}

The ground-truth product information is the ONLY authoritative source of product information. You MUST follow these rules:

1. \textit{Do not question the correctness of the ground-truth product.}

* Treat all provided ground-truth products as correct, relevant, and valid answers to the query. Do not question, verify, or challenge whether a ground-truth product is appropriate for the query. Use the provided ground-truth products directly as the basis for your answer.

2. \textit{Do not invent product information.}

* Do not introduce product names, brands, prices, features, specifications, materials, sizes, colors, functions, availability, ratings, reviews, use cases, or any other product attributes that are not explicitly supported by the provided ground-truth information.

3. \textit{Do not use external knowledge.}

* Even if you know something about a product, brand, category, or product type from your prior knowledge, do NOT use that information unless it is explicitly stated in the provided ground-truth products.

4. \textit{Do not make unsupported inferences.}

* Do not infer a product property merely because it is typical or likely for that type of product.

5. \textit{Do not add information simply to make the answer sound more natural.}

* Natural language generation must not come at the expense of factual grounding.

6. \textit{Use only the provided ground-truth products.}

* Do not recommend, mention, or compare against products that are not included in the ground-truth input.

7. \textit{Every ground-truth product must be mentioned.}

* All the ground-truth products are relevant. You must mention all of them in the answer. If some products are the same by name, combine them into a single mention.

8. \textit{Do not claim facts about the user.}

* The query represents the user's information need, but you should not infer personal preferences, circumstances, budget, location, or intended usage beyond what is explicitly stated in the query.

\textbf{Handling Multiple Products}

If multiple ground-truth products are relevant:

* Mention them naturally rather than mechanically listing all products.

* Distinguish between products using only their explicitly provided information.

* Do not claim one product is ``better'', ``best'', ``cheaper'', ``more suitable'', or ``more highly rated'' unless the provided information explicitly supports such a comparison.

* Do not rank, filter, exclude, or select among the ground-truth products, since all provided products are already considered relevant. Present the relevant options without inventing a ranking.

\textbf{Output Format}

Return exactly the following structure:

Answer: <answer>

Do not output any additional sections, analysis, reasoning, notes, or metadata.
\end{promptbox}

\begin{promptbox}[Prompt for MAG dataset]
You are a grounded natural-language answer generation model in a retrieval-and-generation system.

\textbf{Task}

\textbf{You are given two inputs:}

\textbf{1. User Query}

* A natural-language query representing the user's information need or search intent.

\textbf{2. Ground-Truth Papers}

* A set of one or more ground-truth papers associated with the user query.

* Each paper may contain information such as: paper title, abstract, publication date, venue, journal, conference, citations, field of study, authors, or other paper-related information. If the information is ``nan'', it is unknown and do not use it.

* The ground-truth papers are the ONLY source of paper-related factual information that may be used when generating the answer.

\textbf{Generate an answer for the user query based on the provided information.}

* Answer the query directly and concisely in a natural and conversational way, rather than sounding like a bibliography, database entry, or search result.

* Use the complete paper title exactly as provided in the input whenever a paper is mentioned. Do not abbreviate, truncate, paraphrase, normalize, or replace the paper title.

* Present the papers naturally as part of a coherent answer rather than as a raw list. Do not use bullet points, numbered lists, tables, headings, or any other list-like format.

* Write the entire answer as one continuous paragraph without line breaks or separate sections.

* Use only factual information explicitly provided for the corresponding paper.

\textbf{Grounding and Factuality Requirements}

The ground-truth paper information is the ONLY authoritative source of paper-related information. You MUST follow these rules:

1. \textit{Do not question the correctness of the ground-truth paper.}

* Treat all provided ground-truth papers as correct, relevant, and valid answers to the query. Do not question, verify, or challenge whether a ground-truth paper is appropriate for the query. Use the provided ground-truth papers directly as the basis for your answer.

2. \textit{Do not invent paper information.}

* Do not introduce paper titles, authors, venues, dates, research topics, methods, datasets, experiments, results, contributions, findings, conclusions, or any other paper attributes that are not explicitly supported by the provided ground-truth information.

3. \textit{Do not use external knowledge.}

* Even if you know something about a paper, author, venue, topic, method, dataset, or scientific concept from your prior knowledge, do NOT use that information unless it is explicitly stated in the provided ground-truth papers.

4. \textit{Do not make unsupported inferences.}

* Do not infer a paper's research topic, method, contribution, finding, or relevance merely because it is typical or likely for its title, venue, authors, or research area.

5. \textit{Do not add information simply to make the answer sound more natural.}

* Natural language generation must not come at the expense of factual grounding.

6. \textit{Use only the provided ground-truth papers.}

* Do not recommend, mention, or compare against papers that are not included in the ground-truth input.

7. \textit{Every ground-truth paper must be mentioned.}

* All the ground-truth papers are relevant. You must mention all of them in the answer. If some papers have the same title, combine them into a single mention.

8. \textit{Do not claim facts about the user.}

* The query represents the user's information need or research intent, but you should not infer the user's research background, preferences, expertise, research goals, or intended use beyond what is explicitly stated in the query.

9. \textit{Do not overstate scientific claims.}

* Do not describe a paper as proving, demonstrating, establishing, outperforming, or achieving something unless such a claim is explicitly supported by the provided paper information. Do not convert tentative or qualified statements in the provided paper information into definitive claims.

\textbf{Handling Multiple Papers}

If multiple ground-truth papers are relevant:

* Mention them naturally rather than mechanically listing all papers.

* Distinguish between papers using only their explicitly provided information.

* Do not claim one paper is ``better'', ``best'', ``more relevant'', ``more comprehensive'', ``more effective'', or otherwise superior to another unless the provided paper information explicitly supports such a comparison.

* Do not rank, filter, exclude, or select among the ground-truth papers, since all provided papers are already considered relevant. Present the relevant papers as part of a coherent answer without implying a ranking.

\textbf{Output Format}

Return exactly the following structure:

Answer: <answer>

Do not output any additional sections, analysis, reasoning, notes, or metadata.
\end{promptbox}

\begin{promptbox}[Prompt for PRIME dataset]
You are a grounded natural-language answer generation model in a retrieval-and-generation system.

\textbf{Task}

\textbf{You are given two inputs:}

\textbf{1. User Query}

* A natural-language query representing the user's information need or search intent.

\textbf{2. Ground-Truth Medical Concepts}

* A set of one or more ground-truth medical concepts associated with the user query.

* Each medical concept belongs to one of the following concept types: disease, gene/protein, molecular function, drug, pathway, anatomy, effect/phenotype, biological process, cellular component, or exposure.

* Each medical concept may contain information such as: name, type, source, details, or other concept-related information. If the information is ``nan'', it is unknown and do not use it.

* The ground-truth medical concepts are the ONLY source of medical-concept-related factual information that may be used when generating the answer.

\textbf{Generate an answer for the user query based on the provided information.}

* Answer the query directly and concisely in a natural and conversational way, rather than sounding like a bibliography, database entry, or search result.

* Use the complete medical concept name exactly as provided in the input whenever a medical concept is mentioned. Do not abbreviate, truncate, paraphrase, normalize, translate, or replace the concept name.

* Present the medical concepts naturally as part of a coherent answer rather than as a raw list. Do not use bullet points, numbered lists, tables, headings, or any other list-like format.

* Write the entire answer as one continuous paragraph without line breaks or separate sections.

* Use only factual information explicitly provided for the corresponding medical concept.

\textbf{Grounding and Factuality Requirements}

The ground-truth medical concept information is the ONLY authoritative source of medical-concept-related information. You MUST follow these rules:

1. \textit{Do not question the correctness of the ground-truth medical concept.}

* Treat all provided ground-truth medical concepts as correct, relevant, and valid answers to the query. Do not question, verify, or challenge whether a ground-truth medical concept is appropriate for the query. Use the provided medical concepts directly as the basis for your answer.

2. \textit{Do not invent medical concept information.}

* Do not introduce medical concept names, types, diseases, genes, proteins, drugs, pathways, anatomical structures, phenotypes, biological processes, cellular components, exposures, functions, mechanisms, associations, treatments, symptoms, causes, effects, or any other medical attributes that are not explicitly supported by the ground-truth medical concept information.

3. \textit{Do not use external knowledge.}

* Even if you know something about a disease, gene, protein, drug, pathway, anatomy, phenotype, biological process, cellular component, exposure, or other medical concept from your prior knowledge, do NOT use that information unless it is explicitly stated in the provided ground-truth medical concepts.

4. \textit{Do not make unsupported inferences.}

* Do not infer a medical concept's properties, relationships, biological functions, mechanisms, causes, effects, symptoms, treatments, or associations merely because they are typical or likely for that type of concept.

5. \textit{Do not add information simply to make the answer sound more natural.}

* Natural language generation must not come at the expense of factual grounding.

6. \textit{Use only the provided ground-truth medical concepts.}

* Do not recommend, mention, or compare against medical concepts that are not included in the ground-truth input.

7. \textit{Every ground-truth medical concept must be mentioned.}

* All the ground-truth medical concepts are relevant. You must mention all of them in the answer. If some medical concepts have the same name, combine them into a single mention.

8. \textit{Do not claim facts about the user.}

* The query represents the user's information need or medical information need, but you should not infer the user's medical condition, diagnosis, symptoms, treatment history, health status, preferences, or circumstances beyond what is explicitly stated in the query.

9. \textit{Do not provide unsupported medical advice or conclusions.}

* Do not diagnose the user, recommend a treatment, suggest a medication, assess medical risk, or provide clinical advice unless such information is explicitly requested by the query and directly supported by the provided ground-truth medical concepts.

* Do not transform information about a medical concept into a personalized medical recommendation or clinical conclusion.

10. \textit{Do not overstate medical or biological claims.}

* Do not describe a medical relationship, biological mechanism, causal relationship, therapeutic effect, disease association, or other scientific or medical claim as established unless it is explicitly supported by the provided ground-truth medical concept information.

* Do not convert tentative, qualified, or descriptive statements in the provided medical concept information into definitive claims.

\textbf{Handling Multiple Medical Concepts}

If multiple ground-truth medical concepts are relevant:

* Mention them naturally rather than mechanically listing all medical concepts.

* Distinguish between medical concepts using only their explicitly provided information.

* When medical concepts have different concept types, identify or distinguish their types only when the concept type is explicitly provided and doing so helps answer the query.

* Do not claim one medical concept is ``better'', ``more important'', ``more relevant'', ``more strongly associated'', or otherwise superior to another unless the provided medical concept information explicitly supports such a comparison.

* Do not rank, filter, exclude, or select among the ground-truth medical concepts, since all provided concepts are already considered relevant. Present the relevant medical concepts as part of a coherent answer without implying a ranking.

\textbf{Output Format}

Return exactly the following structure:

Answer: <answer>

Do not output any additional sections, analysis, reasoning, notes, or metadata.
\end{promptbox}

\raggedbottom  %
\subsection{Case Studies}
\label{app:cases}

\begin{resultbox}[Case 1: Amazon (test query 116)]
\raggedright
\textbf{Query $Q$.} ``What are some highly recommended scuba diving weights for experienced divers that would fit well with my recently purchased Gorilla PRO XL waterproof bag?''\\[2pt]
\textbf{Ground-truth entity.} Sea Pearls Vinyl Coated Lace Thru Weights.\par\smallskip
{\footnotesize\textit{Top 3 of each method's final ranking (CARA: RRF; ARK: count-based aggregation):}\par}\vspace{1pt}
{\footnotesize\setlength{\tabcolsep}{3pt}\renewcommand{\arraystretch}{1.12}
\begin{tabular}{@{}c>{\raggedright\arraybackslash}p{0.415\linewidth}>{\raggedright\arraybackslash}p{0.415\linewidth}@{}}
\toprule
Rank & \textbf{ARK} & \textbf{CARA}\\
\midrule
1 & Pocket Weights 10Lb.\ BCD Scuba Weight & \textbf{Sea Pearls Vinyl Coated Lace Thru Weights} (ground truth)\\\addlinespace[3pt]
2 & Scuba Choice Spearfishing \ldots{} Rubber Weight Belt & Arrow Weights 10lb Uncoated Lead Weight \ldots\\\addlinespace[3pt]
3 & Pocket Weights 7Lb.\ BCD Scuba Weight & Pocket Weights 7Lb.\ BCD Scuba Weight\\
\midrule
\multicolumn{3}{@{}l}{\textit{Position of the ground truth in each agent's answer list}}\\
 & none, 5, 15 & 17, 1, 1\\\addlinespace[2pt]
\multicolumn{3}{@{}l}{\textit{Position of the ground truth in the final ranking}}\\
 & 7 & \textbf{1}\\
\bottomrule
\end{tabular}\par}
\end{resultbox}

\medskip
\begin{resultbox}[Case 2: MAG (test query 2057)]
\raggedright
\textbf{Query $Q$.} ``Show me publications on Nanocantilevers that explore magnetism at the nanoscale.''\\[2pt]
\textbf{Ground-truth entity.} Room temperature magnetometry of an individual iron filled carbon nanotube acting as nanocantilever.\par\smallskip
{\footnotesize\textit{Top 3 of each method's final ranking (CARA: RRF; ARK: count-based aggregation):}\par}\vspace{1pt}
{\footnotesize\setlength{\tabcolsep}{3pt}\renewcommand{\arraystretch}{1.12}
\begin{tabular}{@{}c>{\raggedright\arraybackslash}p{0.415\linewidth}>{\raggedright\arraybackslash}p{0.415\linewidth}@{}}
\toprule
Rank & \textbf{ARK} & \textbf{CARA}\\
\midrule
1 & A geometry for optimizing nanoscale magnetic resonance force microscopy & \textbf{Room temperature magnetometry of an individual iron filled carbon nanotube \ldots} (ground truth)\\\addlinespace[3pt]
2 & Magnetic Resonance Force Microscopy Detected Long-Lived Spin \ldots & Scanned probe imaging of nanoscale magnetism \ldots\\\addlinespace[3pt]
3 & Magnetic Resonance Force Microscopy of a Permalloy \ldots & An individual iron nanowire-filled carbon nanotube \ldots\\
\midrule
\multicolumn{3}{@{}l}{\textit{Position of the ground truth in each agent's answer list}}\\
 & none, none, none & 1, 14, 1\\\addlinespace[2pt]
\multicolumn{3}{@{}l}{\textit{Position of the ground truth in the final ranking}}\\
 & not retrieved & \textbf{1}\\
\bottomrule
\end{tabular}\par}
\end{resultbox}

\medskip
\begin{resultbox}[Case 3: Prime (test query 6548)]
\raggedright
\textbf{Query $Q$.} ``Identify genes or proteins known to interact with EPS8L2 and share an association with a common disease.''\\[2pt]
\textbf{Ground-truth entity.} ACTG1.\par\smallskip
{\footnotesize\textit{Top 3 of each method's final ranking (CARA: RRF; ARK: count-based aggregation):}\par}\vspace{1pt}
{\footnotesize\setlength{\tabcolsep}{3pt}\renewcommand{\arraystretch}{1.12}
\begin{tabular}{@{}c>{\raggedright\arraybackslash}p{0.415\linewidth}>{\raggedright\arraybackslash}p{0.415\linewidth}@{}}
\toprule
Rank & \textbf{ARK} & \textbf{CARA}\\
\midrule
1 & HNF4A & \textbf{ACTG1} (ground truth)\\\addlinespace[3pt]
2 & LRRK2 & ROR1\\\addlinespace[3pt]
3 & PTK2 & deafness, autosomal recessive\\
\midrule
\multicolumn{3}{@{}l}{\textit{Position of the ground truth in each agent's answer list}}\\
 & 7, none, 1 & 1, 1, 6\\\addlinespace[2pt]
\multicolumn{3}{@{}l}{\textit{Position of the ground truth in the final ranking}}\\
 & 6 & \textbf{1}\\
\bottomrule
\end{tabular}\par}
\end{resultbox}

\end{document}